\documentclass{article}

\usepackage[final]{tackling_climate_workshop_style}

\usepackage[utf8]{inputenc} 
\usepackage[T1]{fontenc}    
\usepackage{hyperref}       
\usepackage{url}            
\usepackage{booktabs}       
\usepackage{amsfonts}       
\usepackage{nicefrac}       
\usepackage{microtype}      
\usepackage{subcaption}
\usepackage{graphicx}
\usepackage{wrapfig}

\title{Mapping Farmed Landscapes from Remote Sensing}

\author{Michelangelo Conserva \\
	Google Research
	\And
	Alex Wilson \\
	Google Research
	\And
	Charlotte Stanton \\
	Google Research
	\And
	Vishal Batchu \\
	Google Research
	\And
	Varun Gulshan \\
	Google Research
}
\begin{document}

\maketitle

\begin{abstract}
To overcome the critical lack of detailed ecological maps needed for managing agricultural landscapes, we developed \textit{Farmscapes}: the first large-scale, high-resolution map that identifies ecologically vital rural features, including often overlooked elements like hedgerows and stone walls. We achieved high accuracy in mapping key habitats with a deep learning model trained on aerial imagery and expert annotations. As a result, this work enables data-driven planning for habitat restoration, supports the monitoring of key initiatives like the EU Biodiversity Strategy, and lays a foundation for advanced analysis of landscape connectivity.
The dataset is available in Google Earth Engine at \url{https://developers.google.com/earth-engine/datasets/catalog/projects_nature-trace_assets_farmscapes_england_v1_0}.
\end{abstract}

\section{Introduction}

Agricultural systems face increasing pressure to meet the growing demand for food while maintaining ecological balance and mitigating climate impacts \citep{webb2020urgency}.
The intensification of agriculture over recent decades has led to significant environmental challenges, including habitat loss, soil degradation, and declining wildlife populations \citep{landis2017designing}.

In intensively farmed landscapes, natural habitats such as woodlands, grasslands, and hedgerows are often fragmented or removed altogether \citep{krauss2010habitat}.
In this scenario, remaining patches of semi-natural habitats are disconnected from one another, creating isolated islands in a sea of monocultures.
This fragmentation limits species movement, reduces genetic exchange, and diminishes overall biodiversity \citep{wan2018habitat}.
Biodiversity loss, in turn, weakens the resilience of ecosystems, reducing their ability to provide essential services such as pollination, pest control, and climate regulation services that are crucial to sustainable agriculture \citep{lanz2018expansion}.

The global urgency to address biodiversity loss is underscored by international commitments such as the Kunming-Montreal Global Biodiversity Framework, which includes the landmark ``30 by 30'' initiative to protect 30\% of terrestrial and marine ecosystems by 2030 \citep{protectedplanet, cbd2022kunming}.
This ambitious goal has been adopted at a continental level by the EU Biodiversity Strategy for 2030, compelling Member States to enhance habitat protection and restoration, with a particular focus on farmland biodiversity \citep{hermoso2022eu}. 

While these initiatives reflect a growing consensus on the need for large-scale restoration, their success hinges on implementing targeted, cost-efficient interventions \citep{duke2013cost}. A fundamental prerequisite for such strategic action is the availability of accurate maps of rural landscapes, which are essential for guiding governments and local initiatives in prioritising restoration efforts and allocating resources effectively \citep{turner2001landscape, moilanen2005prioritizing}.

Current efforts in high-resolution landscape mapping have shown promise but face significant limitations.
While deep learning techniques are now state-of-the-art for segmenting remote sensing imagery, their application to fine-scaled rural features remains limited.
Studies on hedgerow mapping often suffer from coarse image resolution that misses fine details, misclassification with other woody vegetation, or methodologies that lead to imprecise boundaries \citep{ahlswede2021hedgerow, strnad2023detection, muro5011709hedgerow}.
Research on stone wall mapping has leveraged high-resolution LiDAR data, but the intensive manual labour required for labelling has restricted these efforts to smaller areas \citep{suh2023regional, Trotter19032022}. Furthermore, existing commercial and institutional datasets are often limited by high costs or a lack of methodological transparency, hindering widespread scientific use \citep{hedgeUk, bluesky_national_hedgerow_map}. A critical gap therefore exists for a mapping resource that is simultaneously large-scale, high-resolution, and openly accessible.

To address these limitations, we introduce \textit{Farmscapes}: a large-scale, high-resolution (25 cm) open dataset that identifies ecologically vital rural features across England, with preliminary, unvalidated maps generated for other European countries.
This map was produced using a deep learning model trained on a novel dataset of landscape elements annotations collected for imagery in England.
While quantitatively validated across most of England, the model has also been used to generate illustrative landscape maps for other European countries with similar rural characteristics.
By releasing this extensive dataset publicly, we provide ecologists, policymakers, and local communities in England with a powerful, data-driven tool, while offering a foundation for a future pan-European resource.
To address potential privacy concerns, the initial public release is limited to the hedgerow, stone wall, and woodland data layers.
To the best of our knowledge, Farmscapes is the first resource of its kind, enabling precise, large-scale assessments of ecological status and facilitating the targeted restoration efforts needed to meet ambitious biodiversity goals.

\section{Landscape elements}

Agricultural landscapes are composed of a mosaic of distinct features, which we refer to as landscape elements. These elements, both natural and anthropogenic, define the physical structure of the environment and are fundamental to its ecological function. Their composition and spatial arrangement influence biodiversity, species movement, and the provision of ecosystem services. In this study, we focus on mapping four key elements that are vital to the character and health of European farmland: hedgerows, stone walls, woodland, and the surrounding farmed land.

\textbf{Hedgerows} are linear features composed of shrubs and trees that form boundaries in farmed landscapes. Their structure can vary, ranging from low, intensively managed shrubs to complex, unmanaged lines containing mature trees. This structural diversity is key to their ecological importance.

\textbf{Woodland} refers to areas dominated by trees with a more open canopy and lower density than forests, often featuring a developed understory. These ecosystems are crucial for carbon sequestration, soil stability, and water regulation, and they significantly enhance landscape diversity by acting as transitional zones between open land and dense forests.

\textbf{Stone walls} are man-made linear boundaries constructed from stone. While serving agricultural purposes like enclosing livestock, they also function as important microhabitats. The crevices provide shelter for insects, small animals, and reptiles, and support specialised plant life like mosses and lichens, thereby enhancing local biodiversity.

\textbf{Farmed land} comprises areas actively managed for agricultural production, including the cultivation of crops and the raising of livestock. As a dominant global land use, it is fundamental to food security and includes systems ranging from intensive monocultures to diverse agroforestry. In our study, this class represents the matrix in which other landscape elements are embedded.

\section{Data}

In order to train our deep learning model, we collect 25cm aerial imagery (input) and human annotations (labels). To ensure the accuracy of these annotations, we provided the annotators with 1m LiDAR-derived height maps as a supplementary reference tool.

\textbf{Aerial imagery.}
Identifying fine-scaled features like hedgerows requires imagery with a resolution significantly higher than publicly available sources like Sentinel-2 (10m). To address this, we used proprietary 25cm resolution aerial imagery, captured over England (2018-2022). This high-resolution dataset served as the sole input for our deep learning model.

\textbf{LiDAR measurements.}
While aerial imagery contains visual cues for height, such as shadows and texture, interpreting them consistently can be ambiguous. To establish a highly accurate and unambiguous ground truth, we equipped our annotators with a 1m resolution height map from the UK Environment Agency’s LiDAR Digital Terrain Model (DTM) \citep{eauk_dtm}. This reference layer allowed them to rapidly and confidently distinguish between features of different heights, such as a tall hedgerow and low scrub. The deep learning model was then trained on the aerial imagery to learn the correlation between these visual cues and the LiDAR-verified labels. Note that the LiDAR data was used exclusively for creating the ground truth and was not an input to the model itself.

\begin{figure}[h!]
\centering
\begin{subfigure}{0.32\textwidth}
  \centering
  \includegraphics[width=\textwidth]{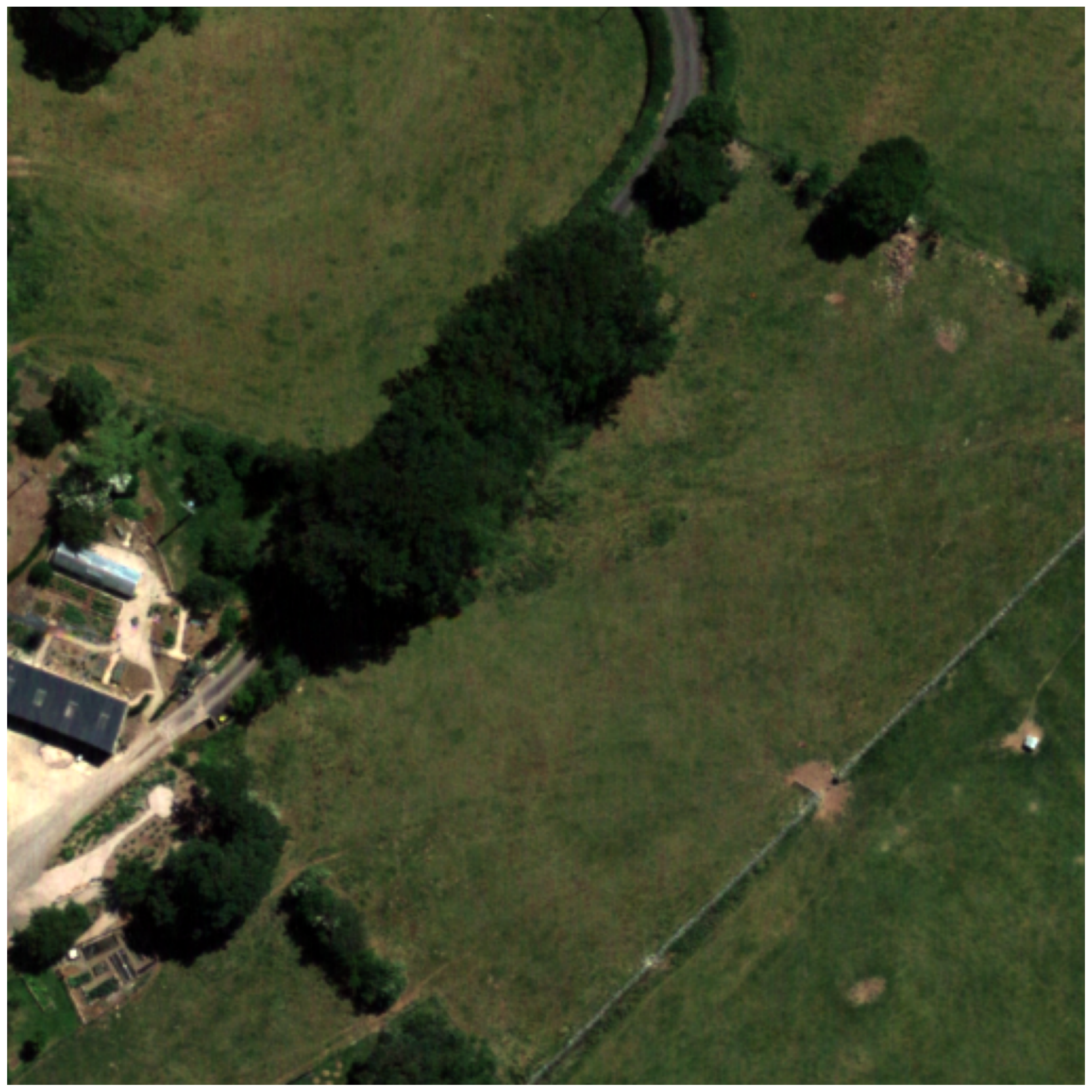}
  \caption{Aerial imagery.}
  \label{fig:annotations_a}
\end{subfigure}
\begin{subfigure}{0.32\textwidth}
  \centering
  \includegraphics[width=\textwidth]{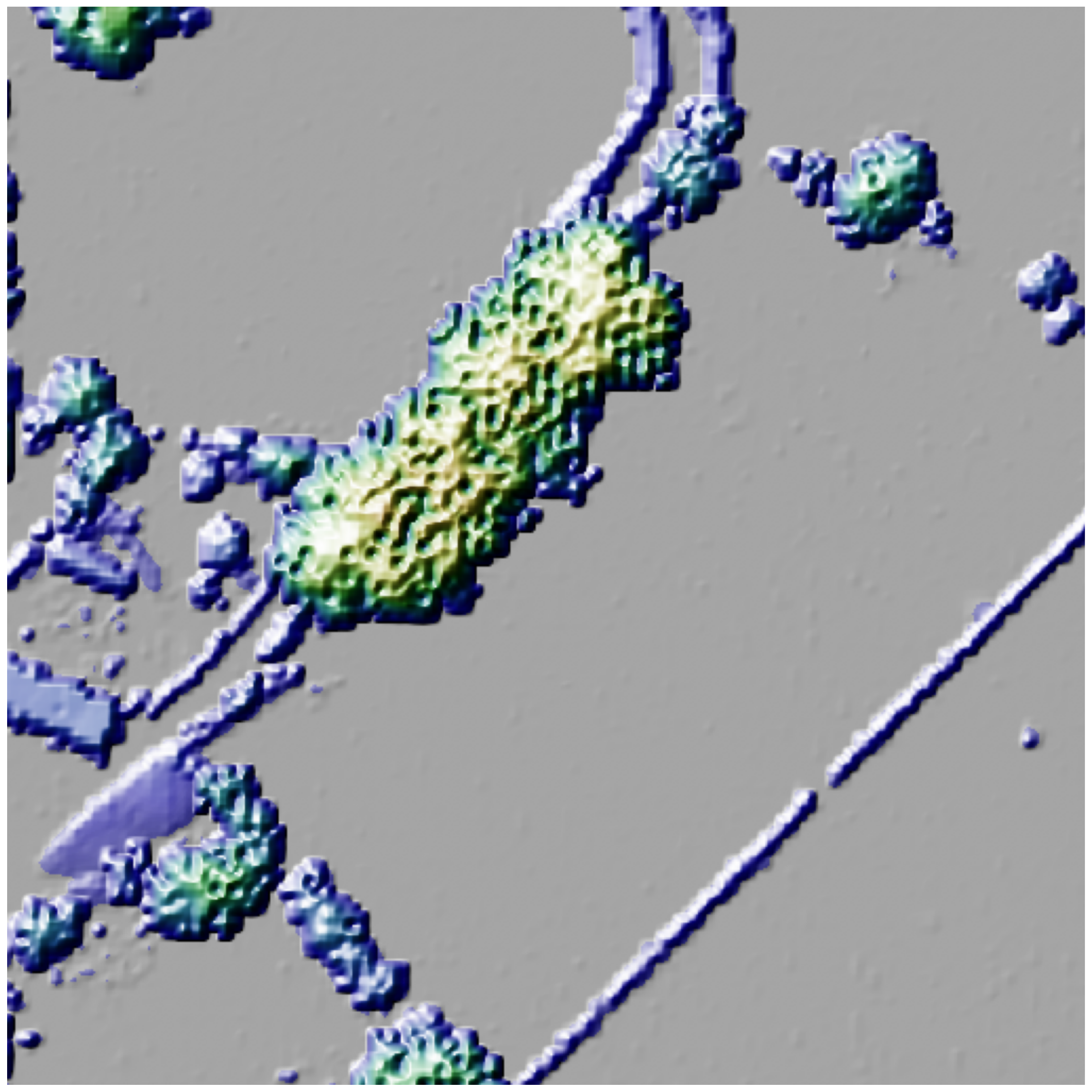}
  \caption{LiDAR-based elevation.}
  \label{fig:annotations_b}
\end{subfigure}
\begin{subfigure}{0.32\textwidth}
  \centering
  \includegraphics[width=\textwidth]{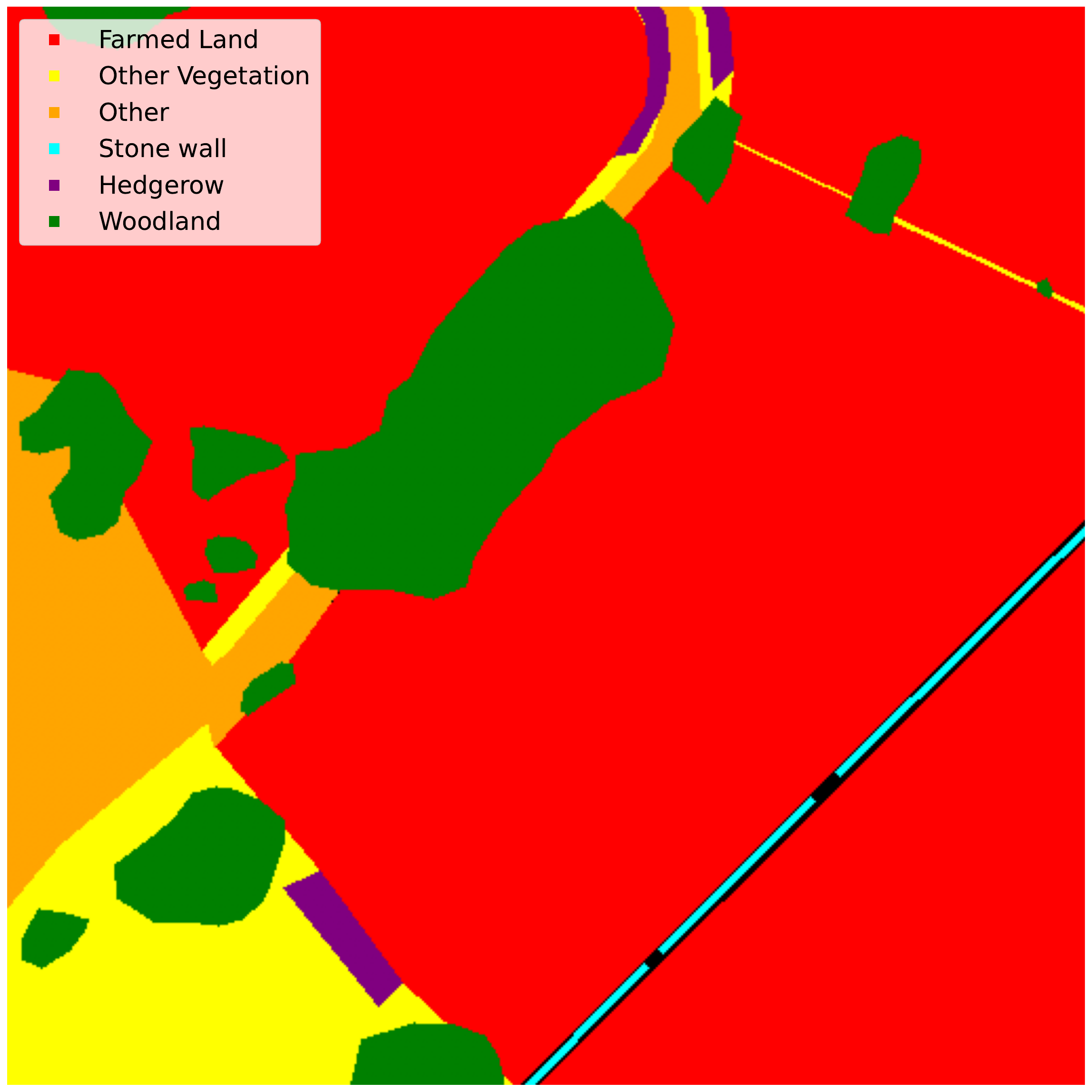}
  \caption{Human annotations.}
  \label{fig:annotations_c}
\end{subfigure}
\caption{Comparison of data sources.}
\label{fig:annotations}
\end{figure}

\textbf{Sampling Strategy}. We created a novel dataset by sampling 942 image tiles from rural England, covering approximately 247 km$^2$. To ensure a representative sample of key features, our process first excluded high-density urban areas and large forests, then used stratified sampling to over-sample regions with a known high density of stone walls. Each tile measures 512m $\times$ 512m, which corresponds to $2048 \times 2048$ pixels at 25cm resolution.

\textbf{Annotation Process}. For annotation, experts used both the aerial imagery (Figure~\ref{fig:annotations_a}) and LiDAR-derived height data (Figure~\ref{fig:annotations_b}) as complementary references. This process resulted in a final annotation mask (Figure~\ref{fig:annotations_c}) labeling seven exhaustive land cover classes: \textit{farmed land}, \textit{hedgerows}, \textit{woodland}, \textit{stone walls}, \textit{other vegetation}, \textit{water}, and \textit{other} (encompassing roads, buildings, etc.).

\textbf{Training targets.}
To model the landscape's vertical structure, where hedgerows or woodland can overlay ground cover, we defined four training targets. A multiclass target was used for mutually exclusive ground classes (\textit{farmed land}, \textit{other vegetation}, \textit{other}), while three separate binary targets were used to identify features that can co-exist within the same pixel space (\textit{hedgerows}, \textit{woodland}, \textit{stone walls}). 
The distribution of training targets is shown in Table~\ref{tab:classes_counts}.

\begin{table}[h!]
    \centering
    \caption{Percentage of classes of the training targets.}
    \setlength{\tabcolsep}{1pt}
    \resizebox{\textwidth}{!}{%
    \begin{tabular}[t]{l@{\hspace{-0.2cm}}r}
        \toprule
         & \textbf{Ground} \\
        \midrule
        Farmed land & 67\% \\
        Vegetation & 27\% \\
        Other & 3\% \\
    \end{tabular}
    \begin{tabular}[t]{l@{\hspace{-0.25cm}}r}
        \toprule
         & \textbf{Hedgerow} \\
        \midrule
        Non hedgerow & 99\% \\
        Hedgerow & 1\%
    \end{tabular}
    \begin{tabular}[t]{l@{\hspace{-0.25cm}}r}
        \toprule
         & \textbf{Stone wall} \\
        \midrule
        Non stone wall & 99\% \\
        Stone wall & 1\%
    \end{tabular}
    \begin{tabular}[t]{l@{\hspace{-0.25cm}}r}
        \toprule
         & \textbf{Woodland} \\
        \midrule
        Non woodland & 89\% \\
        Woodland & 11\%
    \end{tabular}
    }
    \label{tab:classes_counts}
\end{table}

\textbf{Deep learning model.}
We use a SETR-PUP segmentation transformer \citep{zheng2021rethinking}, which uses a Vision Transformer (ViT) encoder and separate convolutional decoders for each of the four targets.
The encoder was pretrained on $\sim100$ million global aerial images using a masked autoencoder approach \citep{MAE_He_2022_CVPR}. To improve generalisation, we applied extensive data augmentation, including random flips, rotations, color jitter, resizing, and cutout \citep{devries2017improved}.
The model was trained on 512$\times$512 pixel (128m × 128m) random crops.

\section{Results}

We evaluated our model's performance on the hold-out test set using f1-score, precision, and recall. Classification thresholds were selected to maximise the f1-score on the validation set. F1-scores are presented in Table~\ref{tab:test-metrics}. Precision and recall are presented in the appendix (see Table~\ref{tab:test-metrics-app}).
While these scores establish a strong baseline, direct quantitative benchmarking against state-of-the-art methods is precluded by our use of proprietary 25cm imagery, a step that was essential to resolve fine-scale features like stone walls.

The model demonstrates solid performance on broad landscape features. For major classes like \textbf{Woodland} and \textbf{Farmed land}, it achieved f1-scores of $96 \pm 1$ and $95 \pm 1$, respectively.
This indicates that the model reliably identifies the primary matrix of the agricultural landscape.
The segmentation of fine-scale linear features presented a greater challenge.
Nevertheless, the model achieves an f1-score of $72 \pm 1$ for \textbf{Hedgerows}.
Performance on \textbf{Stone walls} was lower, with an f1-score of $60 \pm 1$, reflecting the difficulty of identifying these features, which are often just a few pixels wide and can be obscured by vegetation.
These results represent a significant advancement in mapping these critical, yet difficult-to-detect, ecological corridors.

\begin{table}[h!]
    \centering
    \caption{Test metrics with average and standard deviation over twelve seeds.}
    \begin{tabular}{lrrrrrrrr}
    \toprule
    & \multicolumn{3}{c}{Ground}& \multicolumn{3}{c}{Above ground} \\
    \cmidrule{2-4} \cmidrule{5-7}
    & Vegetation & Farmed land & Other & Stone walls & Hedgerows & Woodland \\
    \midrule
    f1-score & $84 \pm 1$ & $95 \pm 1$ & $81 \pm 3$ & $60 \pm 1$ & $72 \pm 1$ & $96 \pm 1$ \\
    \bottomrule
    \end{tabular}
    \label{tab:test-metrics}
\end{table}

\section{Conclusion}

In this study, we have successfully demonstrated the power of combining high-resolution aerial imagery with a deep learning approach to produce \textit{Farmscapes}: the first large-scale, open-access map of key rural landscape elements. Our model, which leverages a Vision Transformer architecture, has proven highly effective at segmenting the agricultural matrix, achieving excellent performance in mapping broad features like woodland and farmed land. More importantly, it marks a significant advancement in the automated identification of fine-scale linear features, delivering strong results for hedgerows and a promising baseline for the notoriously challenging task of mapping stone walls. By creating and publicly releasing this dataset, we provide a foundational tool for ecologists, land managers, and policymakers to support data-driven conservation, monitor progress towards ambitious goals like the EU Biodiversity Strategy, and enable new research into landscape connectivity.

This study is subject to four key limitations. First, our use of proprietary 25cm aerial imagery, while essential for resolving fine-scale features like hedgerows and stone walls, precludes direct quantitative benchmarking against state-of-the-art methods. Second, model performance is fundamentally constrained by annotation quality; despite our use of LiDAR to maximize accuracy, the manual labeling process is subject to human error, which can introduce noise. Third, this map represents a static snapshot from 2018-2022 imagery. Monitoring landscape change, a critical need for tracking policy effectiveness, is not yet possible and remains important future work. Finally, quantitative validation was conducted exclusively on data from England, meaning the map’s reliability outside this region remains unquantified; any application to other geographies must be considered preliminary.

Building on this foundation, future work will proceed along three main avenues. First, we aim to refine the model's accuracy, challenging classes like stone walls, by exploring other architectures and complementary data sources. Second, we plan to broaden the map's scope, geographically and temporally. This involves collecting new annotated data from diverse European landscapes to improve generalization and adapting the model to work with lower-resolution imagery to enable historical change analysis. Finally, our ultimate goal is to translate this detailed map into actionable ecological insights. By conducting a comprehensive network analysis of landscape connectivity, we can identify critical corridors, fragmentation points, and priority areas for restoration, transforming \textit{Farmscapes} into a dynamic tool for strategic environmental planning.

\bibliographystyle{apalike}
\bibliography{bibliography}

\clearpage
\appendix

\section{Annotation methodology and guidelines}

\begin{figure}[h!]
\centering
\begin{subfigure}{0.32\textwidth}
  \centering
  \includegraphics[width=\linewidth]{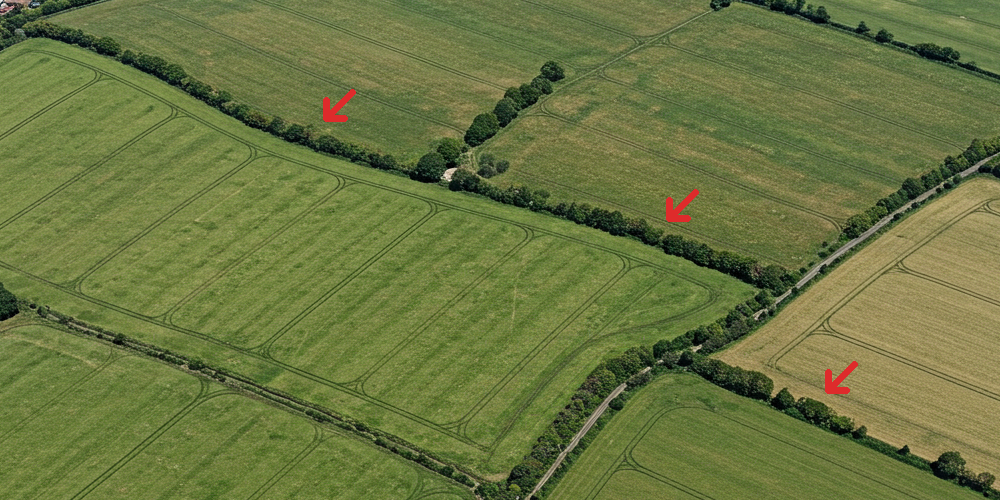}
\end{subfigure}
\hfill
\begin{subfigure}{0.32\textwidth}
  \centering
  \includegraphics[width=\linewidth]{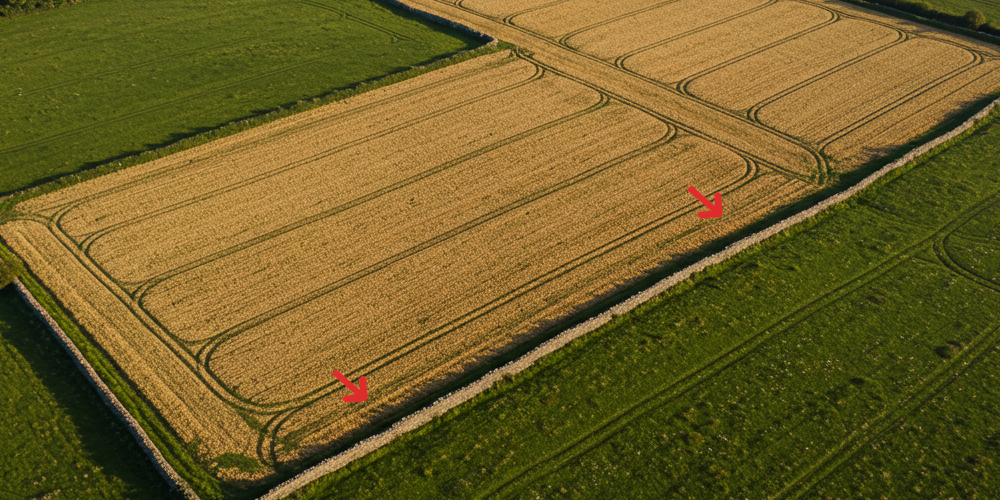}
\end{subfigure}
\hfill
\begin{subfigure}{0.32\textwidth}
  \centering
  \includegraphics[width=\linewidth]{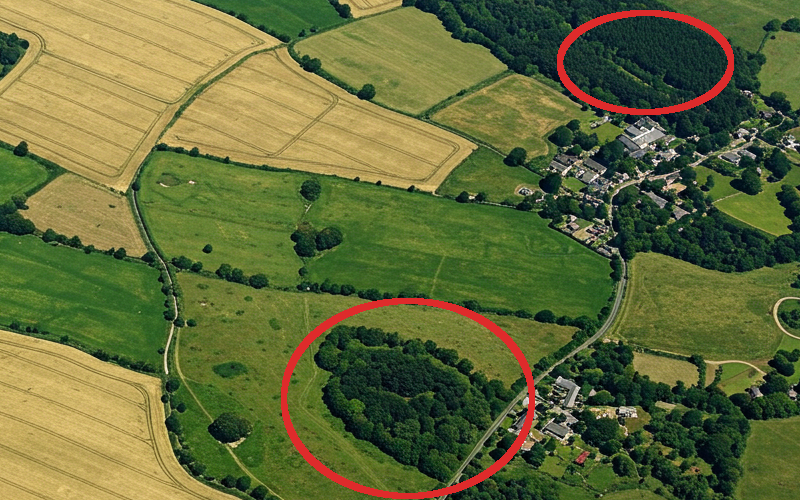}
\end{subfigure}
\caption{Typical English landscape elements, hedgerows (left), stone walls (center), and woodland (right).}
\label{fig:landscapes}
\end{figure}

\textbf{Landscapes elements.}
Figure~\ref{fig:landscapes} presents visual examples of the primary landscape elements targeted in this study: hedgerows, stone walls, and woodland.

\begin{figure}[h!]
\centering
\begin{subfigure}{0.3\textwidth}
  \centering
  \includegraphics[height=6.5cm]{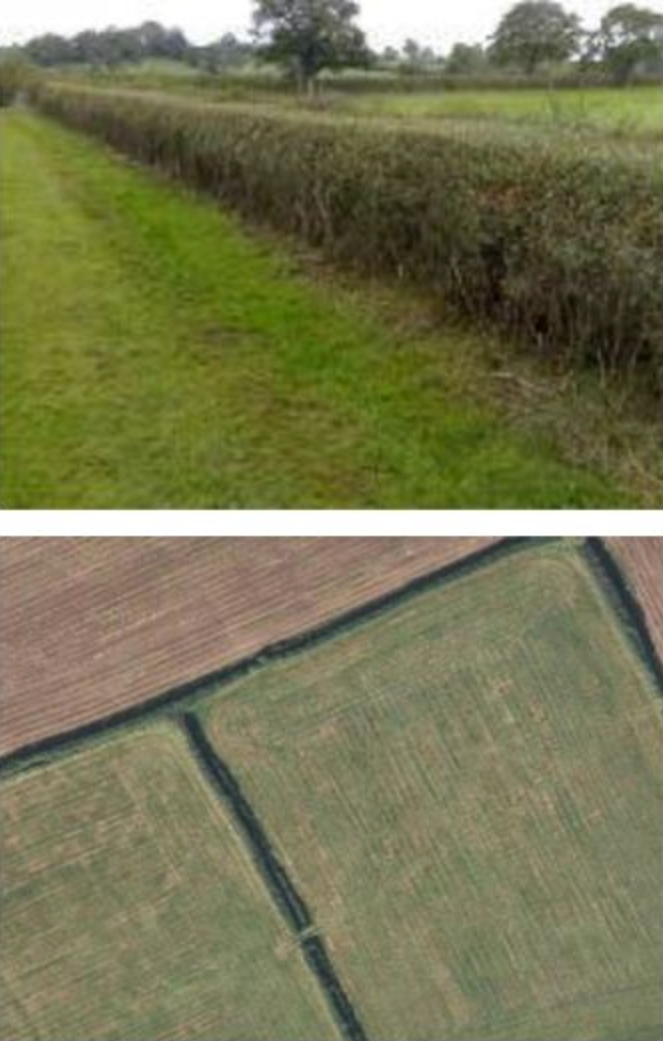}
  \caption{Hedgerow Type 1.}
  \label{fig:hedge_1}
\end{subfigure}
\begin{subfigure}{0.3\textwidth}
  \centering
  \includegraphics[height=6.5cm]{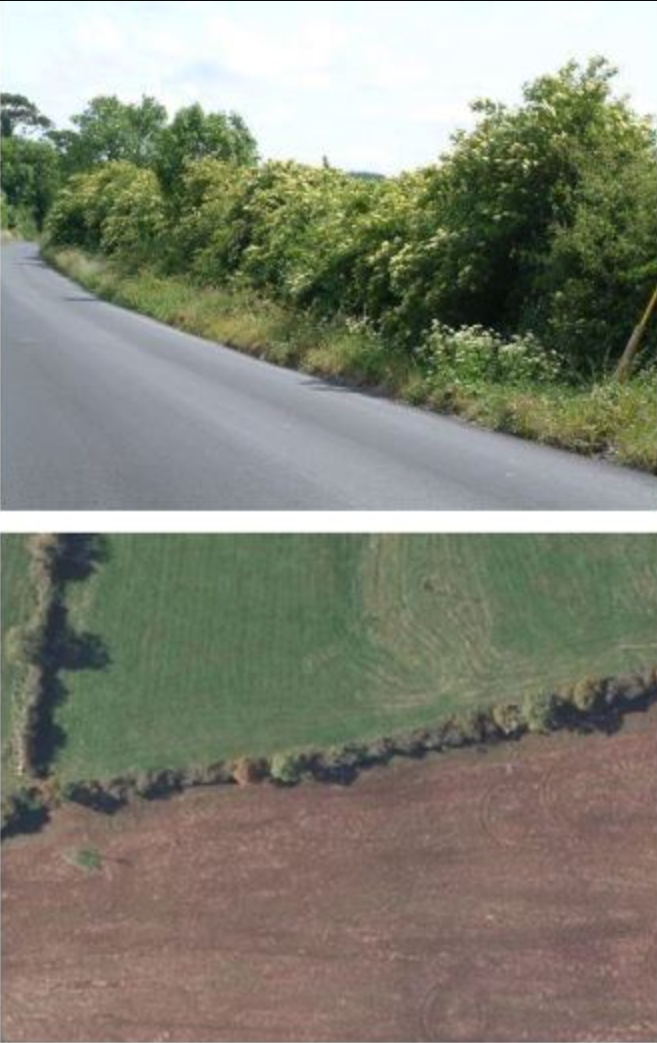}
  \caption{Hedgerow Type 2.}
  \label{fig:hedge_2}
\end{subfigure}
\begin{subfigure}{0.3\textwidth}
  \centering
  \includegraphics[height=6.5cm]{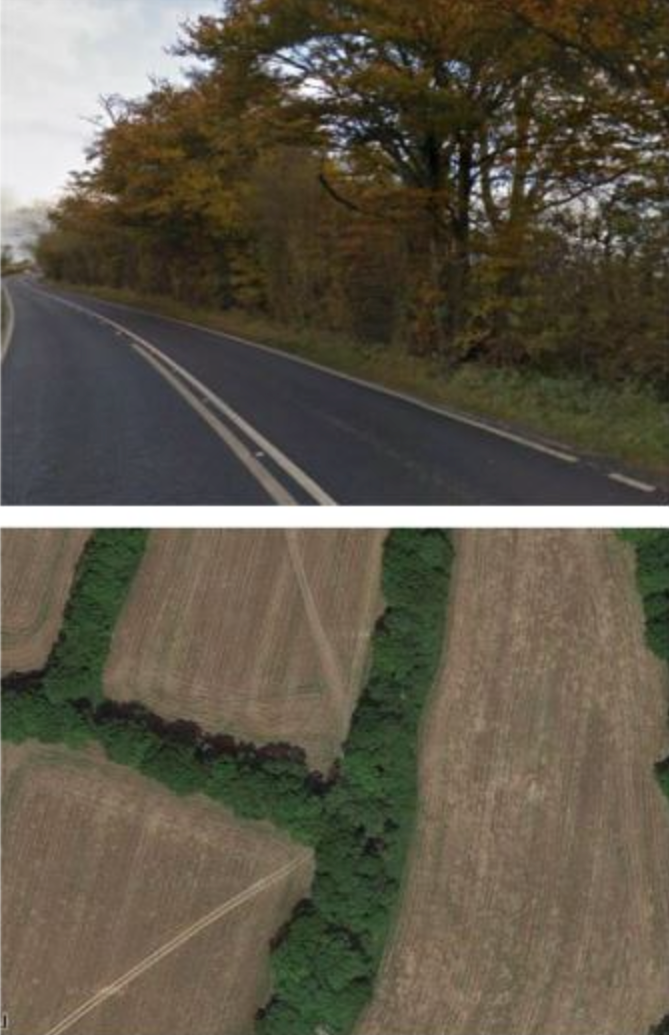}
  \caption{Hedgerow Type 3.}
  \label{fig:hedge_3}
\end{subfigure}
\caption{Types of hedgerows.}
\end{figure}

\textbf{Hedgerow Annotation Guidelines.}
To ensure consistent labelling of hedgerows, which exhibit significant structural diversity, our annotation protocol follows the typology proposed by \citet{neumann2016compositional}. This framework defines three distinct hedgerow types based on their structure and management intensity. The key rule for our annotation process was to map these types to our final land cover classes as follows:
\begin{itemize}
\item \textbf{Type 1:} A low-lying, intensively managed hedge composed of shrubs and lacking mature trees (Figure~\ref{fig:hedge_1}). According to \citet{neumann2016compositional}, these typically reach up to 1.5 meters in height with an average width of 2.5 meters.
\item \textbf{Type 2:} A less managed, taller hedgerow containing small trees or tall shrubs (Figure~\ref{fig:hedge_2}). These features are structurally more complex than Type 1, with an average width of 7 meters.
\item \textbf{Type 3:} A hedgerow that includes mature trees, often appearing from an aerial perspective as a narrow, linear strip of woodland (Figure~\ref{fig:hedge_3}). Due to its structural similarity to forest edges, this type was classified as \textit{Woodland} to maintain a clear distinction based on the presence of a mature tree canopy.
\end{itemize}

\begin{figure}[h!]
\centering
\begin{subfigure}{0.4\textwidth}
  \centering
  \includegraphics[height=6cm]{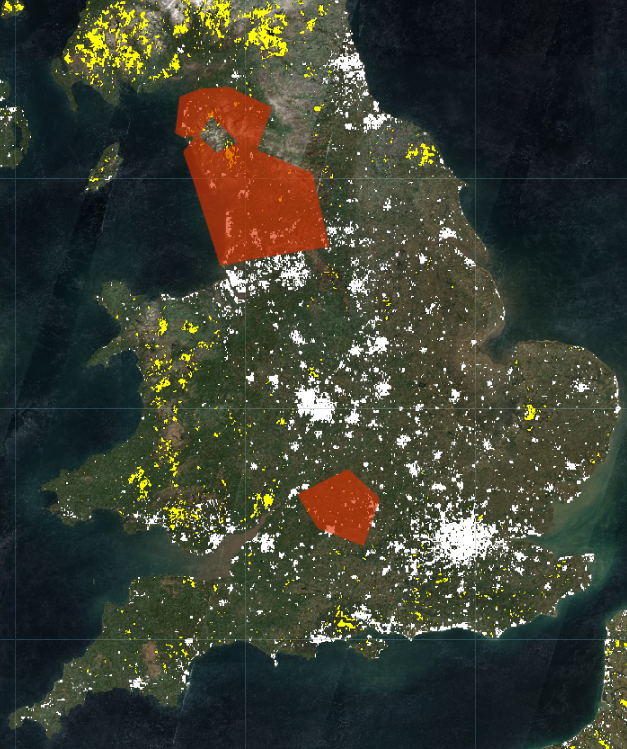}
  \caption{High-density urban areas (white), forests (yellow), and areas with higher stone wall density (red).}
  \label{fig:sampling_a}
\end{subfigure}
\hspace{2cm}
\begin{subfigure}{0.4\textwidth}
  \centering
  \includegraphics[height=6cm]{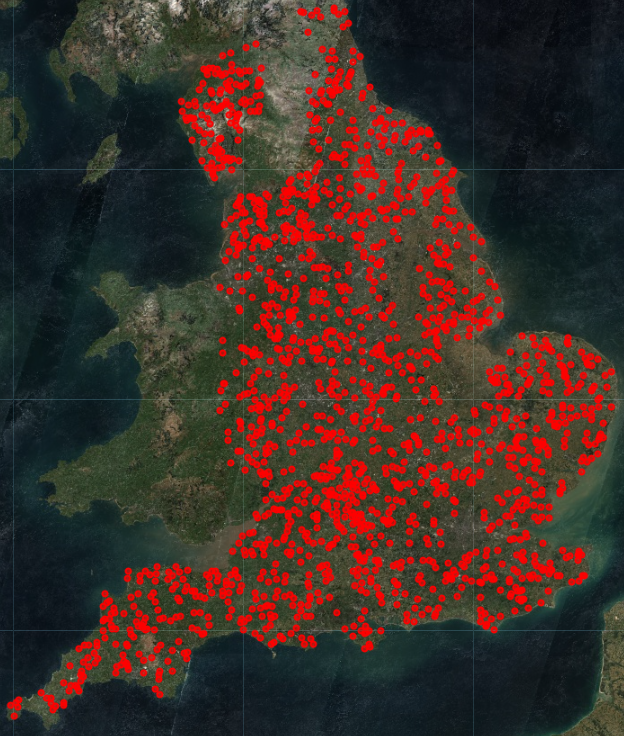}
  \caption{Locations sampled for annotation (red) superimposed on areas with available aerial imagery.}
  \label{fig:sampling_b}
\end{subfigure}
\caption{Sampling strategy illustration.}
  \label{fig:sampling}
\end{figure}

\textbf{Sampling Strategy.}
The 942 image tiles for annotation were sampled from across England using a stratified approach designed to ensure representative coverage of agricultural landscapes while oversampling for the rare stone wall class. The process was as follows:
\begin{enumerate}
\item \textbf{Exclusion Masking:} We first excluded non-target areas. High-density urban regions were masked using the World Settlement Footprint \citep{Marconcini2020}, and large forest blocks were masked using the Global Forest Change dataset \citep{hansen2013high}. This focused the sampling on the rural matrix.
\item \textbf{Stratification:} The remaining area was divided into two strata to ensure adequate representation of stone walls (Figure \ref{fig:sampling_a}). Stratum 1 consisted of regions with a known high density of stone walls (e.g. Cotswold and the Lake District), while Stratum 2 comprised all other eligible rural areas.
\item \textbf{Random Sampling:} We then performed stratified random sampling, allocating 15\% of the total tiles to Stratum 1 and the remaining 85\% to Stratum 2. Figure \ref{fig:sampling_b} shows the final distribution of sampled locations.
\end{enumerate}
This strategy ensures the dataset captures a wide variety of rural landscapes while providing sufficient examples of the less common stone wall class for effective model training.

\section{Related works}

In the context of high-resolution landscape mapping, hedgerow mapping is one of the most studied applications \citep{pirbasti2024hedgerows}.
In Germany, \citet{ahlswede2021hedgerow} trained a DeepLab v3 \citep{chen2017rethinking} model using 1-meter resolution imagery from the IKONOS mission \citep{dial2003ikonos}. Labels were manually digitized hedgerow polygons provided by the Bavarian State Office for the Environment. Although this study improved upon the use of coarser imagery, it was limited by the resolution, which may miss finer details such as small gaps in hedgerows critical for ecological analysis. Additionally, the focus solely on hedgerows excluded other important landscape features.
\citet{strnad2023detection} employed a U-Net model trained on aerial photography of Slovenia, with a high spatial resolution of 25 cm. Labels for this study were derived from a reference layer based on LiDAR point clouds collected in 2014 and processed to exclude buildings. While this approach allowed for scaling without manual labelling, it was limited to identifying woody vegetation and could not specifically distinguish hedgerows.
More recently, \citet{muro5011709hedgerow} used multi-temporal PlanetScope satellite imagery with a 3-meter resolution to map hedgerows across Germany using a U-Net architecture. Labels were sourced from a dataset created by the Schleswig-Holstein State Office for Agriculture, Environment, and Rural Areas, combining digital terrain models and high-resolution imagery. However, this approach buffered hedgerow labels by five meters, potentially leading to over-segmentation and the loss of fine details.
In addition to academic research, institutional efforts have also played a role in advancing hedgerow mapping. The UK Centre for Ecology and Hydrology offers a dataset of linear hedgerows, providing a valuable resource for ecological studies \citep{hedgeUk}. Similarly, Bluesky \citep{bluesky_national_hedgerow_map} offers a map of hedgerows and trees across the UK, which incorporates detailed volumetric information.
While these datasets offer significant potential for large-scale analyses, their high cost and lack of transparency regarding methodologies and validation processes limit broader accessibility and utility.

In parallel, mapping stone walls has also benefited from advances in remote sensing and deep learning.
\citet{suh2023regional} mapped stone walls in the Northeastern USA using U-Net, and \citet{diakogiannis2020resunet} models with high-resolution airborne LiDAR data. The model input consisted of LiDAR-derived hillshades and slope maps, with labels created through manual digitization of stone walls from LiDAR data, supplemented by aerial imagery, Google Street View, and field verification. Similarly, \citet{Trotter19032022} focused on updating Denmark's stone wall registry using a U-Net model applied to LiDAR-derived terrain data. The model input included a Digital Terrain Model (DTM), Height Above Terrain (HAT), and a Sobel-filtered DTM, all at a 40 cm resolution. Labels were generated from a stone wall dataset provided by the Danish Ministry of Culture, validated and corrected using the DTM. While both approaches demonstrate the effectiveness of using LiDAR data to detect stone walls, the high manual labor required to create the labels limits the scale of the datasets.

\section{Training details}

\begin{figure}[h!]
    \centering
    \includegraphics[width=0.7\linewidth]{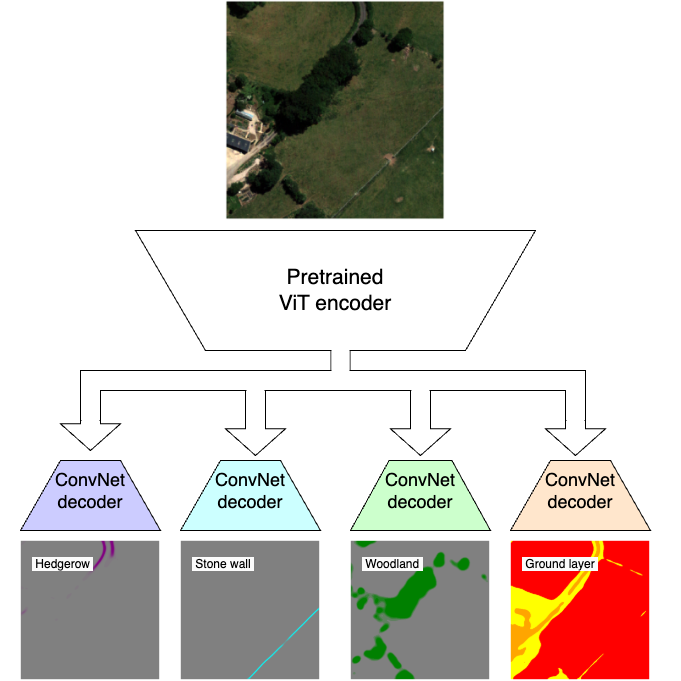}
    \caption{Model architecture.}
\end{figure}

\paragraph{Hyperparameter optimization.}
The 942 tiles of the collected datasets are partitioned into 742 for training, 100 for validation, and 100 for testing.
To accurately assess model performance, validation and test tiles were carefully hand-selected. This selection process prioritized tiles with the clearest annotations and a representative number of instances for rarer classes like stone walls and hedgerows.
We optimized the training of hyperparameters by sweeping over weight decay and learning rate. Performance for each hyperparameter combination was averaged across three random seeds evaluated on the validation set.

\section{Detailed performance metrics}

\begin{table}
    \centering
    \caption{Test metrics with average and standard deviation over twelve seeds.}
    \begin{tabular}{lrrrrrrrr}
    \toprule
    & \multicolumn{3}{c}{Ground}& \multicolumn{3}{c}{Above ground} \\
    \cmidrule{2-4} \cmidrule{5-7}
    & Vegetation & Farmed land & Other & Stone walls & Hedgerows & Woodland \\
    \midrule
    f1-score & $84 \pm 1$ & $95 \pm 1$ & $81 \pm 3$ & $60 \pm 1$ & $72 \pm 1$ & $96 \pm 1$ \\
    recall & $90 \pm 1$ & $91 \pm 1$ & $90 \pm 1$ & $53 \pm 1$ & $70 \pm 1$ & $92 \pm 1$ \\
    precision & $85 \pm 1$ & $96 \pm 1$ & $82 \pm 4$ & $58 \pm 1$ & $73 \pm 1$ & $90 \pm 1$ \\
    \bottomrule
    \end{tabular}
    \label{tab:test-metrics-app}
\end{table}

\paragraph{Ground Layer.}
For the \textit{Vegetation} class, the model achieves an f1-score of $84 \pm 1$, a recall of $90 \pm 1$, and a precision of $85 \pm 1$. This indicates that the model can accurately identify general vegetation cover with high consistency and completeness.  The \textit{Farmed land} class exhibits even higher performance, with an f1-score of $95 \pm 1$, a recall of $91 \pm 1$, and a precision of $96 \pm 1$, demonstrating the model's strong ability to delineate agricultural areas. The \textit{Other} class, encompassing a variety of non-vegetated, non-farmed land covers, achieves an f1-score of $81 \pm 3$, a recall of $90 \pm 1$, and a precision of $82 \pm 4$. While the precision shows some variability, the high recall indicates that the model captures most of this diverse class.

\paragraph{Stone Walls.}
The model achieves an f1-score of $60 \pm 1$ for stone walls.  The recall is $53 \pm 1$, and the precision is $58 \pm 1$. This suggests that while the model has some ability to identify stone walls, there's room for improvement, particularly in capturing all instances (recall).  The relatively balanced precision and recall indicate that the model is neither overly prone to false positives nor false negatives, but rather struggles with consistent detection of these narrow features.

\paragraph{Hedgerows.}
For hedgerows, the model exhibits an f1-score of $72 \pm 1$, a recall of $70 \pm 1$, and a precision of $73 \pm 1$. These results demonstrate a good ability to identify hedgerows, with a reasonable balance between correctly identifying them (precision) and capturing all instances (recall).  The performance on hedgerows is notably higher than that on stone walls, likely due to their typically larger size and more distinct visual characteristics.

\paragraph{Woodland.}
The model performs exceptionally well on the \textit{Woodland} class, achieving an f1-score of $96 \pm 1$, a recall of $92 \pm 1$, and a precision of $90 \pm 1$. This indicates that the model can very accurately and consistently identify woodland areas. The high f1-score, combined with high recall and precision, demonstrates the model's strong capability in delineating this important land cover type.

\clearpage
\section{Qualitative examples}

\begin{figure}[h!]
    \centering
    \includegraphics[width=0.9\linewidth]{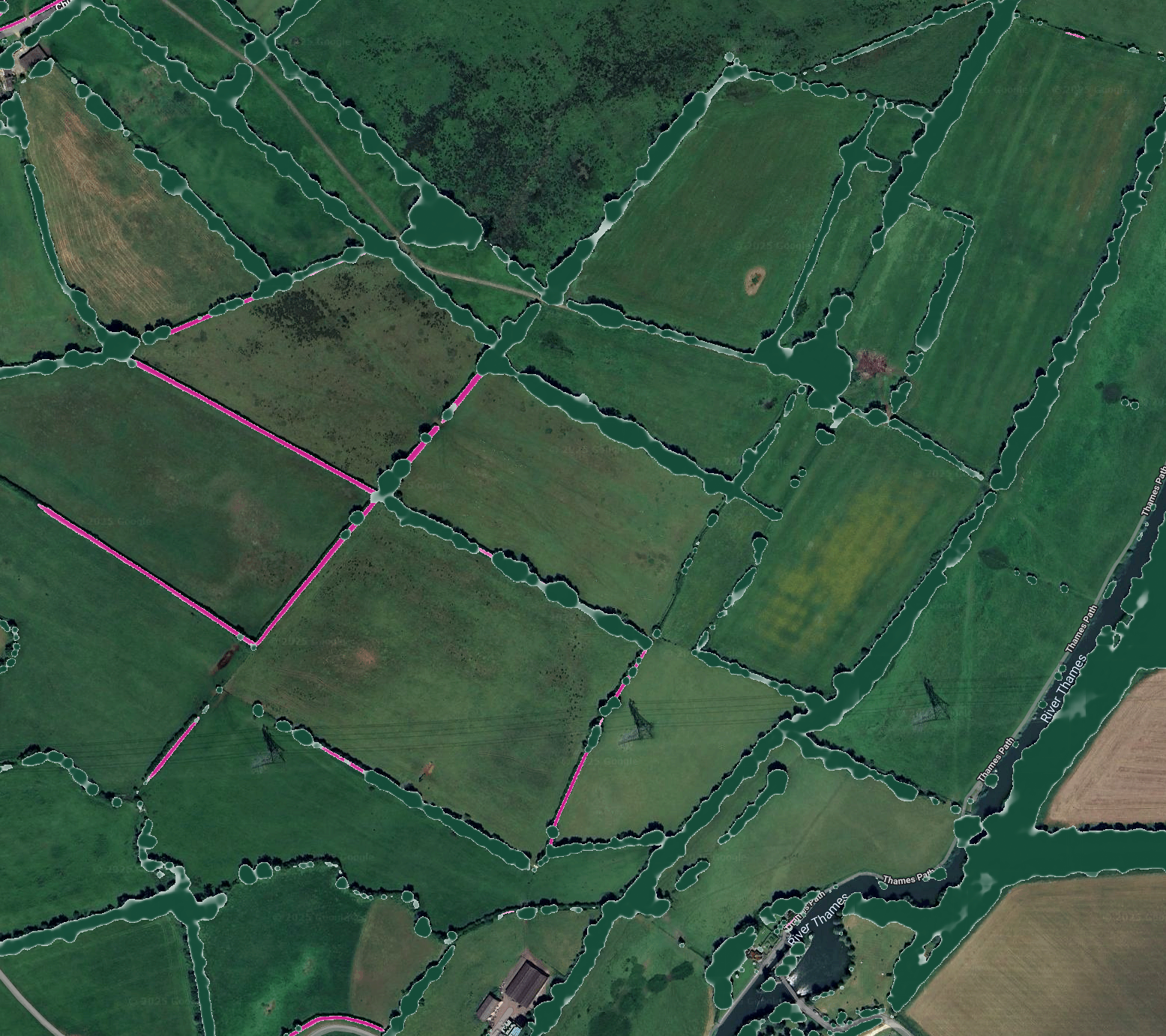}
    \caption{English landscape.}
\end{figure}

\begin{figure}[h!]
    \centering
    \includegraphics[width=0.9\linewidth]{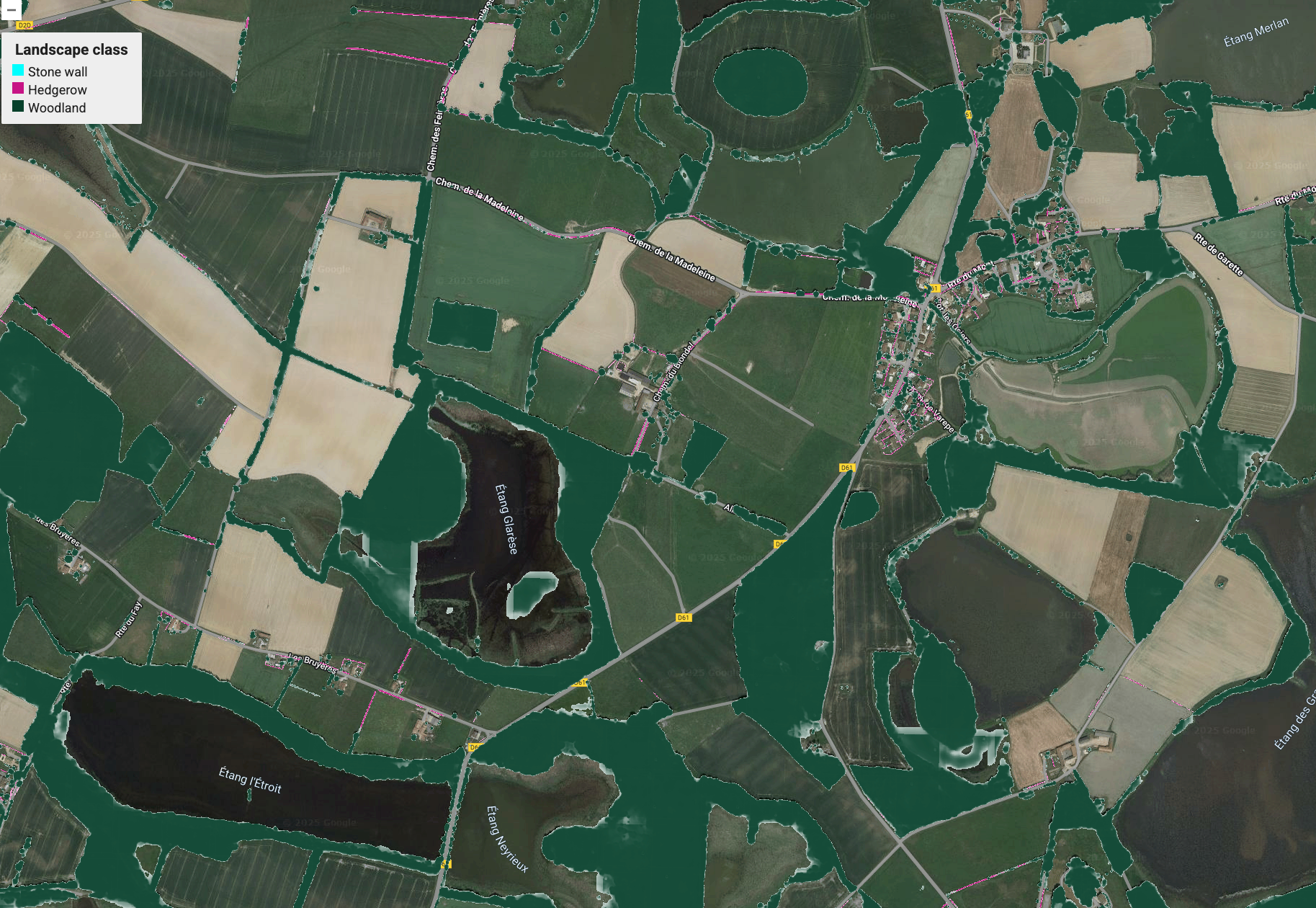}
    \caption{France landscape.}
\end{figure}

\begin{figure}[h!]
    \centering
    \includegraphics[width=0.9\linewidth]{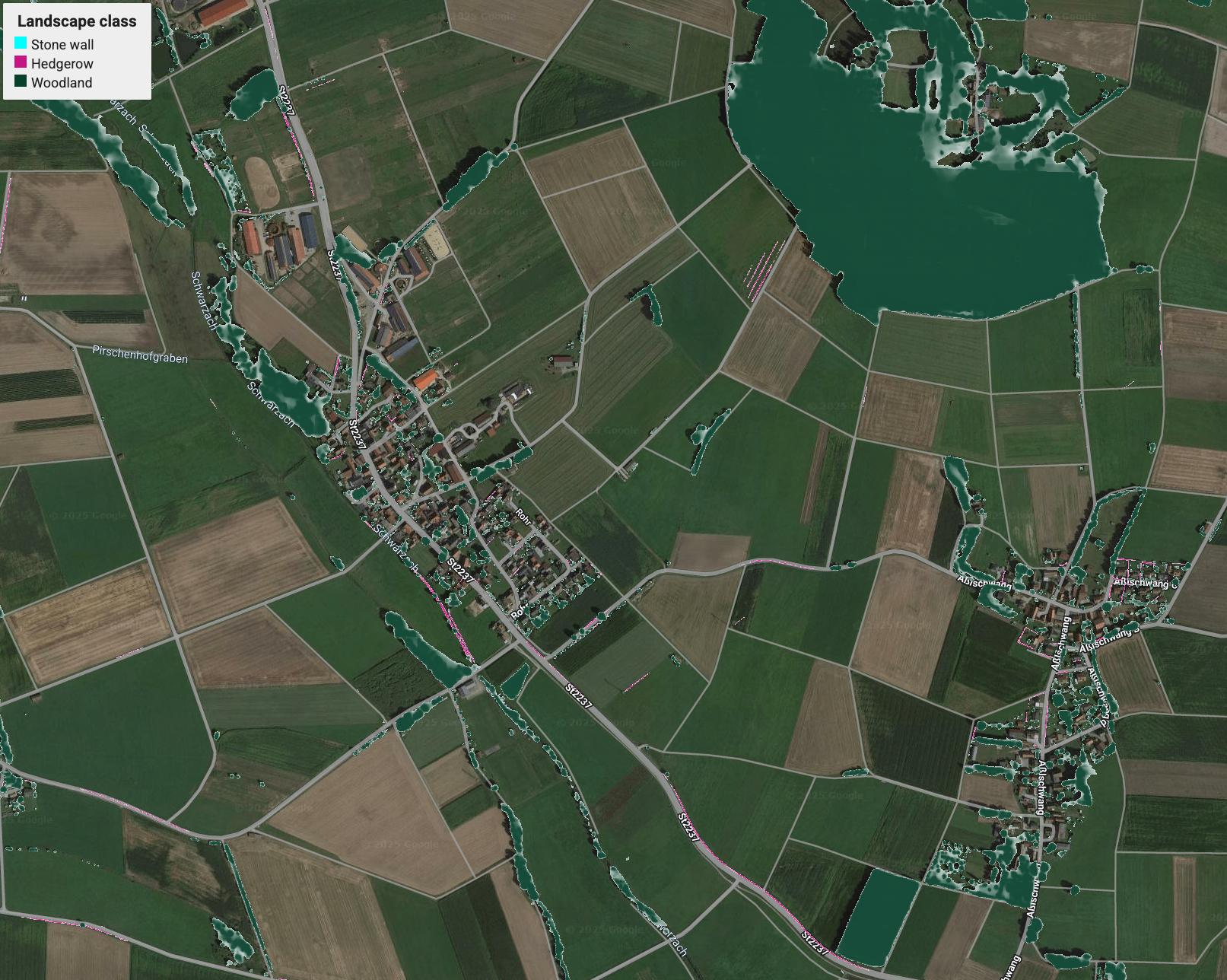}
    \caption{Germany landscape.}
\end{figure}

\begin{figure}[h!]
    \centering
    \includegraphics[width=0.9\linewidth]{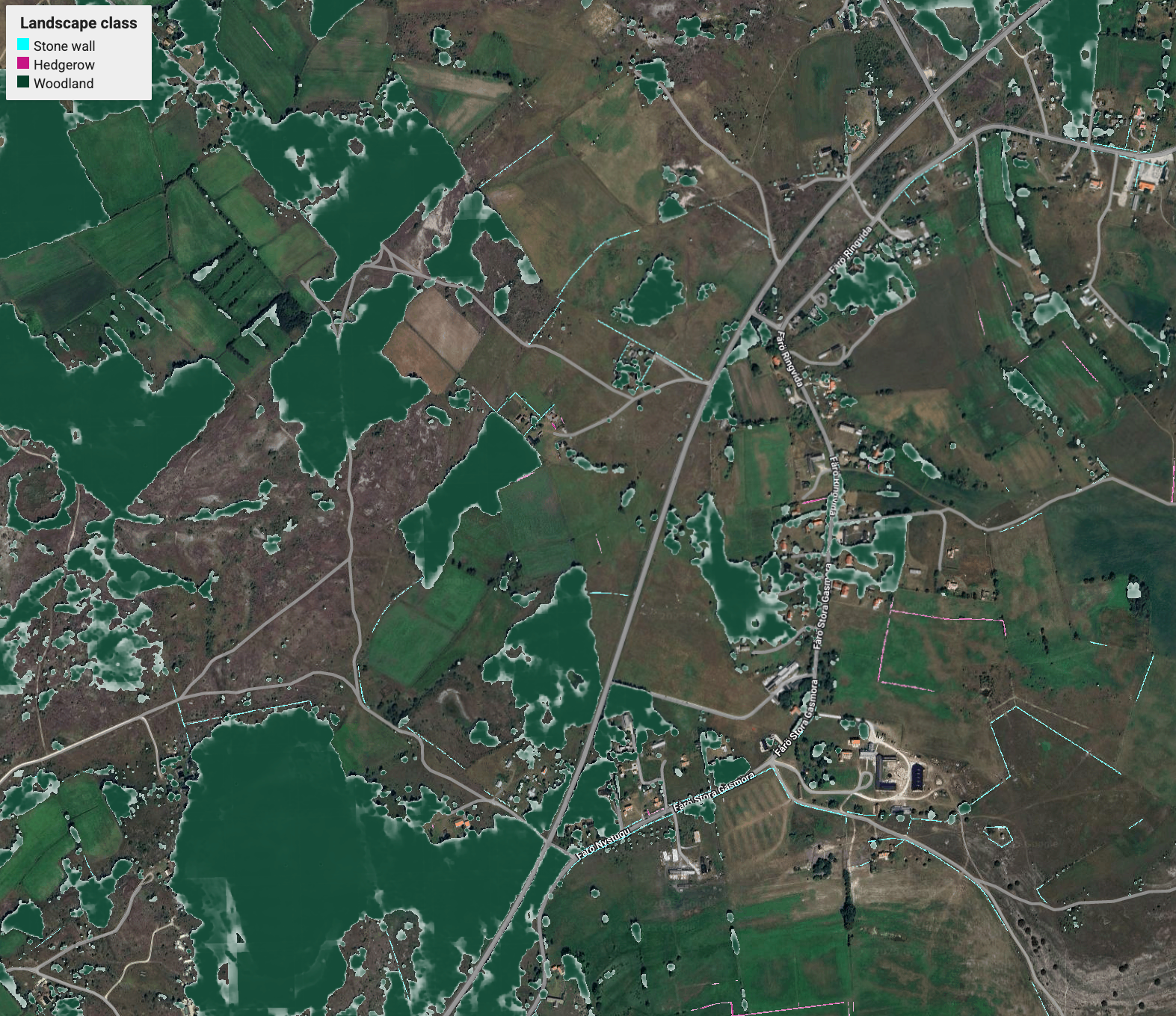}
    \caption{Sweden landscape.}
\end{figure}

\begin{figure}[h!]
    \centering
    \includegraphics[width=0.85\linewidth]{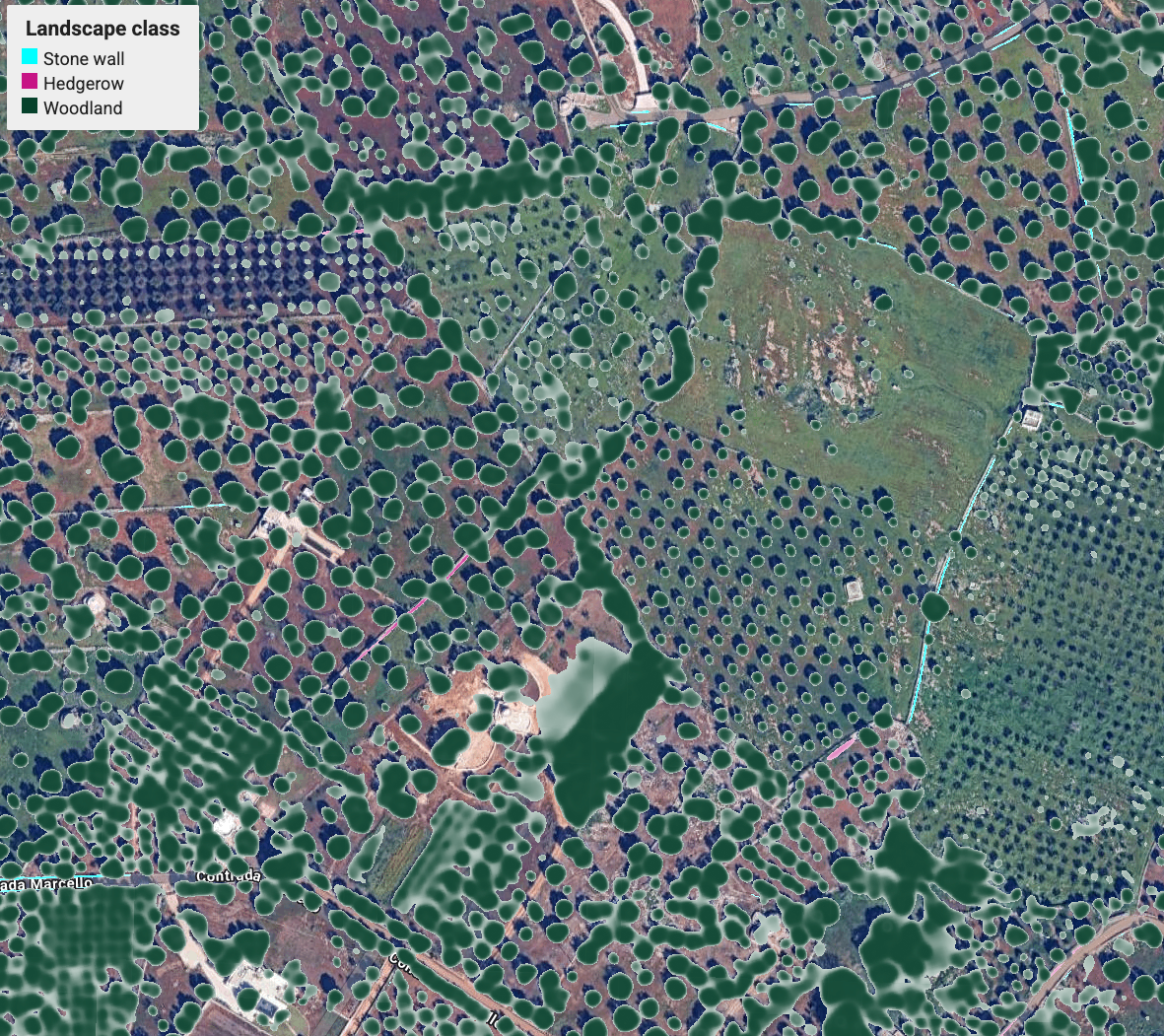}
    \caption{Italy landscape.}
\end{figure}

\begin{figure}[h!]
    \centering
    \includegraphics[width=0.88\linewidth]{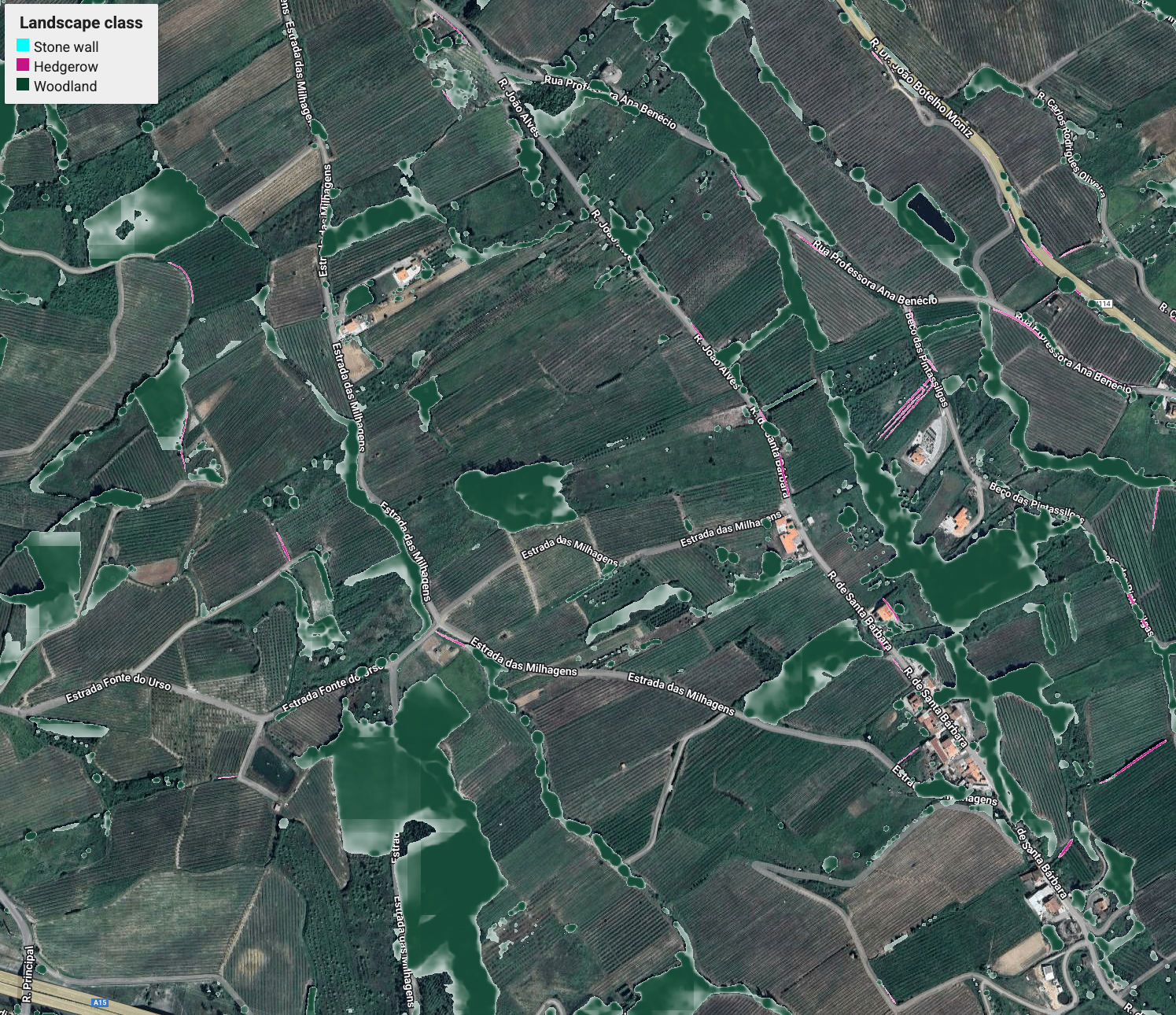}
    \caption{Portugal landscape.}
\end{figure}

\end{document}